\documentclass[acmsmall]{acmart}
\usepackage[utf8]{inputenc}
\usepackage{graphics}
\usepackage{graphicx}
\usepackage{amsthm}
\usepackage{amsmath}
\usepackage{blindtext}
\usepackage{geometry}
\usepackage[compact]{titlesec}
\usepackage{tikz}
\usepackage{mathtools, nccmath}
\usepackage{extdash}
\usepackage{hyperref}
\usepackage{subcaption}
\AtBeginDocument{%
  \providecommand\BibTeX{{%
    \normalfont B\kern-0.5em{\scshape i\kern-0.25em b}\kern-0.8em\TeX}}}

\setcopyright{none}

\renewcommand\footnotetextcopyrightpermission[1]{} 
\title{High Performance Out-of-sample Embedding Techniques for Multidimensional Scaling}
\date{August 2021}

\begin{document}

\author{Samudra Herath}
\affiliation{%
  \institution{University of Adelaide}
  \streetaddress{North Terrace}
  \city{Adelaide}
  \state{South Australia}
  \country{Australia}
  \postcode{5000}
}
\email{samudra.herath@adelaide.edu.au}
\orcid{1234-5678-9012}

\author{Matthew Roughan}
\affiliation{%
  \institution{University of Adelaide}
  \streetaddress{North Terrace}
  \city{Adelaide}
  \state{South Australia}
  \country{Australia}
  \postcode{5000}
}
\email{matthew.roughan@adelaide.edu.au}

\author{Gary Glonek}
\affiliation{%
  \institution{University of Adelaide}
  \streetaddress{North Terrace}
  \city{Adelaide}
  \state{South Australia}
  \country{Australia}
  \postcode{5000}
}
\email{gary.glonek@adelaide.edu.au}

\renewcommand{\shortauthors}{Herath, et al.}

\begin{abstract}

The recent rapid growth of the dimension of many datasets means that many approaches to dimension reduction (DR) have gained significant attention. High-performance DR algorithms are required to make data analysis feasible for big and fast data sets. However, many traditional DR techniques are challenged by truly large data sets. In particular multidimensional scaling (MDS) does not scale well. MDS is a popular group of DR techniques because it can perform DR on data where the only input is a dissimilarity function. However, common approaches are at least quadratic in memory and computation and, hence, prohibitive for large-scale data. 

We propose an out-of-sample embedding (OSE) solution to extend the MDS algorithm for large-scale data utilising the embedding of only a subset of the given data. We present two OSE techniques: the first based on an optimisation approach and the second based on a neural network model. With a minor trade-off in the approximation, the out-of-sample techniques can process large-scale data with reasonable computation and memory requirements. While both methods perform well, the neural network model outperforms the optimisation approach of the OSE solution in terms of efficiency. OSE has the dual benefit that it allows fast DR on streaming datasets as well as static databases. 

\end{abstract}

\maketitle

\section{Introduction}

Once a dataset has been transformed, it is common to need to incorporate new data. This is known as the out-of-sample embedding (OSE) problem. Data points outside of the original sample (used to construct the embedding) must be added, and it is highly desirable that this does not invoke a complete re-computation both because of the computation cost and because we might wish to preserve properties of the original embedding for visualisation or other purposes. 

For instance, in sensor networks, this problem occurs when one seeks to automatically map the sensors' locations given the pairwise distances between them and then infer the locations of new targets as and when they appear~\cite{Wang}. 


Among DR techniques, we examine multidimensional scaling (MDS)~\cite{Kruskal1964}. This is a popular DR technique that approximates points from a high-dimensional space into a lower-dimensional Euclidean space. We focus on MDS in this work because MDS can be used to analyse data defined only by a dissimilarity matrix. This does not even have to be strictly a distance metric, nor does the original space need to be Euclidean. Hence, MDS applies to many data and has applications across many disciplines, including clinical psychology, sociology, marketing, ecology, and biology~\cite{Saeed}. 

The least-squares multidimensional scaling (LSMDS) algorithm is more robust to strongly non-Euclidean data \cite{herath2020simulating}, so we focus on this approach. However, its computation scales as $O(N^2)$ for $N$ data points and is thus impractical for large datasets. 

In addition, many DR techniques, such as LSMDS, do not offer an explicit function to embed out-of-sample objects in an existing space without altering it. Given that the MDS algorithm is $O(N^2)$, re-computation for each new sample is out of the question. 

The provision of a fast OSE technique solves both problems. New points can be easily added, and this makes it possible to build a larger embedding using a subset of landmark points. Thus a high-performance OSE technique extends the LSMDS algorithm to address two main concerns:

\begin{enumerate}
    \item It helps scale the LSMDS algorithm for large-scale data. 
    \item It provides a means querying or adding out-of-sample data into an existing lower-dimensional space.  
\end{enumerate}

Our OSE solution uses only a fraction of the existing configuration to map new points. We call these ``landmarks" and keep them fixed. The mapping of a new object is found by minimising the errors in distances to landmarks (noting, however, that this optimisation is subtly different from that of the original LSMDS). We show here that this works well for mapping names into a Euclidean space. 

We also propose an OSE technique based on an artificial neural network (ANN). The ANN is trained using the distances to the landmarks from the input data and their relevant coordinate values in the existing lower-dimensional space created by MDS. Out-of-sample data is then mapped into this space by the ANN. The training may be slow, but once the ANN is trained, the OSE is fast, so the cost of training is amortised over many OSEs.

Both approaches use landmarks to reduce the computational overhead of dissimilarity calculations. However, the experimental results suggested that the ANN model is on average $ 3.8\times 10^{3}$ times faster than the optimisation approach (with its optimal parameter setting) for predicting a coordinate of a new data point, and both techniques exhibit similar accuracy. Thus the ANN model is the most efficient OSE technique for large-scale LSMDS.  

For a given large-scale data set consisting of $N$ objects, we first apply LSMDS for a selected subset of $L \ll N$ landmarks, with a complexity of $O(L^2)$. These $L$ data points are embedded in a $K$ dimensional space, and we call this a configuration space. We can map the remaining $M=N-L$ objects to the existing configuration applying the proposed OSE approaches with only $O(L M)$ computations. Therefore our approach dramatically reduces the computational complexity.

Similarly, mapping a new object to the above configuration requires only the distances between landmarks and the new object. When mapping a new object, the OSE approaches use landmarks to reduce the complexity of $O(N)$ computations to $O(L)$. Moreover, in the ANN, the computations required are much more efficient than the original LSMDS optimisation.  

In summary, our primary contributions are:
\begin{itemize}

    \item We formally introduce the OSE problem for the LSMDS algorithm. We explain the use of OSE solutions for handling large-scale input data in LSMDS. 
    \item We propose an OSE technique for LSMDS using an optimisation approach.
    \item We propose an ANN model that finds the mapping of a new object faster than the optimisation-based solution. 

\end{itemize}
The combination makes MDS much more practical for large data or online streaming data. 

\section{Problem Formulation}

In this section, we formulate our problem using the relevant definitions and some key concepts. The proposed OSE techniques extend the capabilities of LSMDS. They allow efficient embedding of large-scale data in a lower-dimensional space and mapping new data points that were not in the training sample in an existing configuration.

Assume $R$ is a collection of objects, $\delta$ measures the distances between $R$ objects, $X$ represents the coordinates matrix for the $R$ objects in the Euclidean space, and $d$ measures the distances between coordinates. We define mapping from a metric or non-metric space $(R,\delta)$ into a Euclidean space $(X,d)$ as $\phi : R \to X$. In this paper $(R,\delta)$ can be a finite metric or non-metric space (\textit{i.e.,} $R$ is a finite set) and $(X,d)$ will always be a Euclidean space. 

\newtheorem{mydef}{Problem Statement}
\begin{mydef}
Multidimensional scaling: For a given metric or non-metric space, find a $\phi$ that minimises distortion, stress, or a similar error metric between $(R,\delta)$ and $(X,d)$.
\end{mydef}

\subsection{Least-squares Multidimensional Scaling}

The standard LSMDS method minimises the raw stress of a configuration. The input to the method is a dissimilarity matrix \mbox{$\Delta_N=[\delta_{ij}]$}, given $\delta_{ij}$ is the distance between the indices of two data points $i$ and $j$ in a high-dimensional space. For a given dissimilarity matrix~\mbox{$\Delta_N$}, a configuration matrix \mbox{$\mathbf{X}=[x_1,x_2,..,x_N]$}, where $x_i \in \mathbb{R}^K$ is constructed by minimising 

\begin{equation} \label{eq:1}
\sigma_{raw}(\mathbf{X})={\sum_{i,j=1}^{n}\Big(d_{ij}(\mathbf{X})-\delta_{ij}\Big)^2}.
\end{equation} 

The Euclidean distance between the $i^{th}$ and $j^{th}$ points in the $K$-dimensional space is given  by~$d_{ij}$. One can use squared distances or a normalised raw stress criterion as well. We prefer the normalised stress ($\sigma$) in our experiments since it is popular and theoretically justified~\cite{Bae:2010:DRV:1851476.1851501}. The normalised stress is obtained by \begin{math}
  \sigma = \sqrt{\sigma_{raw}(\mathbf{X})/\delta_{ij}^2 }
\end{math}.

LSMDS is usually implemented by applying iterative gradient descent on the stress function. Another implementation of LSMDS based on the SMACOF (stress majorization of a complicated function) algorithm has been used in many MDS applications. We applied the gradient descent algorithm for the iterative minimisation of stress in our implementation, whereas \mbox{De Leeuw \textit{et al.}~\cite{Jan2009}} used majorization to minimise stress in SMACOF.

\subsection{Out-of-sample Embedding Problem}

Many dimension reduction methods such as MDS do not offer an explicit OSE function. Hence, mapping new data into a lower-dimensional space without extensively reconstructing the entire configuration from the augment dataset is difficult. Running the MDS algorithm from the beginning using all the data (old data and new data) is not a suitable approach for large-scale multidimensional datasets.

The OSE approaches map additional points to a previously constructed configuration. Suppose we have a configuration of $N$ points in a \mbox{$K$-dimensional} Euclidean space obtained by applying LSMDS to a set of $N$ objects. Let $y$ be an out-of-sample object, with measured pairwise dissimilarities from each of the original $N$ objects. The OSE problem is to embed the new $y$ object into the existing configuration of \mbox{$K$-dimensional} Euclidean space. 

\begin{mydef}
Out-of-sample embedding: 
Finding the coordinates (location) of a new object $y$, in a high dimensional space $(R,\delta)$ by mapping it to an existing configuration space $(X,d)$.
\end{mydef}

We discuss how to use OSE techniques to scale the LSMDS algorithm for large-scale data in Section~4. 

\medskip

\noindent Several comparison methods such as edit distance, a.k.a. Levenshtein distance, Jaro distance, and q-gram distance are found in the domain of strings~\cite{Loo2014}. In this work, we mainly used the Levenshtein distance to measure the similarity between string values. It calculates the minimum number of character insertions, deletions, and replacements necessary to transform a string $s_1$ into a string~$s_2$. Minkowski metrics based on $L^p$ norms, ${\parallel x \parallel}_p = (\sum |x_i|^p )^{1/p}$, with $p \geqslant 1$ are popular in the metric domains. We used the most common Minkowski metric, Euclidean distances $d_E(p=2)$ in metric space calculations.

\section{Related Work}

The OSE problem has been actively researched in recent years to improve dimension reduction (DR) algorithms. Beningo~\emph{et al.}~\cite{Bengio2003} provides a unified framework of out-of-sample extensions for different DR algorithms such as Local linear embedding (LLE), Isomap, Laplacian Eigenmaps, MDS and Spectral Clustering. Many of these algorithms use eigendecomposition to obtain a lower-dimensional embedding of the data on a non-linear manifold. The proposed solutions use the Nystrom sampling (a technique that speeds up kernel method computations) to train a model for out-of-sample points without recomputing eigenvectors. Hence, it is only applicable to classical MDS. Furthermore, these non-parametric out-of-sample extensions have scalability limitations since the complexity linearly increases with the size of the datasets.

In contrast, Trosset and Priebe~\cite{Trosset2008} extended the underlying optimisation problem to resolve the OSE problem with eigenmaps constructed by classical MDS. They solve the optimisation problem numerically, considering the least-squares approximation of the pairwise inner products of the existing space. We use a similar concept to introduce an embedding approach for LSMDS that directly approximates the dissimilarities with distances rather than pairwise inner products.

Anderson and Robinson~\cite{Anderson2003GeneralizedDA} have also proposed an approach for placing a new observation into a configuration space based on inter-point distances. The method uses classical MDS to produce a configuration space of the data and discriminant analysis to test group differences. Both methods proposed by Trosset and Priebe~\cite{Trosset2008}, and Anderson and Robinson~\cite{Anderson2003GeneralizedDA} have produced identical results when embedding a single point to an existing configuration. However, these methods calculate all the distances between a new point and existing points in the original space for mapping it into the existing configuration and are not suitable for large-scale MDS. In contrast, our method only calculates distances to a fixed set of landmarks when mapping a new object in an existing space created by LSMDS. 

Gower~\cite{Gower1968AddingAP} added a single observation to an existing configuration by adding a dimension, which alters the configuration space. In contrast, we are interested in embedding a new object into an existing configuration without changing it because we aim to embed many out-of-sample objects.  

Embedding new points in a previously configured embedding space using LSMDS has been studied earlier. Bae \textit{et al.}~\cite{Bae:2010:DRV:1851476.1851501} proposed an interpolation approach similar to k-nearest neighbour classification that solves the out-of-sample problem of LSMDS. Their approach extends the SMACOF implementation of the LSMDS algorithm. Following similar ideas, we propose OSE methods for LMDS for its gradient descent based implementation. However, there are significant differences between the two approaches.


\begin{itemize}

    \item The I-MDS method works well with large-scale data, and the input data is usually Euclidean distances. In contrast, our solutions work well for large-scale MDS applications that process non-Euclidean input data. 
    
    \item I-MDS uses a subsample of the existing objects to find the mapping of the new data. For each new object, it calculates the relevant nearest neighbours and maps them in the existing space by solving an optimisation problem. There are two notable limitations; firstly, the method depends on the nearest neighbours (NNs). Thus, selecting many NNs for each new object reduces the efficiency. Secondly, k-NN search only works for data in a metric space, e.g., input data are points in a Euclidean space. Hence, I-MDS is not applicable to non-metric input data. These limitations reduce the generalisability of the I-MDS algorithm. In contrast, our methods can take non-metric dissimilarities as inputs when handling large-scale LSMDS problems.
    
\end{itemize}

The OSE problem has been appeared and addressed in various contexts. Some of them are bioinformatics \cite{Dong2002,Bae:2010:DRV:1851476.1851501}, sensor networking \cite{Virtual_L}, high performance computing \cite{Bae:2010:DRV:1851476.1851501} and image processing~\cite{MITO201125}.

ANNs have been used in modelling manifold embedding due to their robust learning framework in recent years. Jansen \textit{et al.}~\cite{JANSEN20171} have used modern deep learning to perform out-of-sample mapping in graph embedding. Their experiments suggested that deep neural network models outperform the Nystrom methods. In \cite{pmlr-v5-maaten09a}, t-SNE (t-distributed stochastic neighbour embedding) was combined with deep feed-forward networks to obtain a highly non-linear embedding function. Bunte \textit{et al.}~\cite{Bunte} formalised well-known non-parametric DR techniques based on cost optimisation to define explicit mapping functions that facilitate out-of-sample extensions. ANNs and parameterised non-linear functions are used in the formalisation process. These methods can handle large-scale data since the mapping functions learn from a small random subset of the data. However, the current settings of this process only work for Euclidean data.

By looking at the existing methods of OSE for MDS, we found three main limitations:
\begin{itemize}
    \item Many of them only provide solutions for classical MDS, not LSMDS and as a byproduct focus on embedding Euclidean data. 
    \item Most of these existing solutions do not address the large-scale MDS problems. Instead, the focus is on investigating the technical difficulty that arises by inserting a small number of additional points into previously constructed configurations.
    \item All the methods are tested only on Euclidean data. Hence, the out-of-sample methods for non-metric MDS applications have not been studied well.  
\end{itemize}

We will address these limitations by proposing an OSE approach for LSMDS, which is computationally efficient and works with non-metric data. In the following section, we will explain our methods in detail.


\section{Methods for Large-scale LSMDS}

One way to improve the speed of LSMDS mapping without losing the embedding quality is to consider only a fraction of the input data initially. Hence, the first step is to apply LSMDS only on a subset of all input objects. In the second step, the remaining data can be added to the existing configuration as out-of-sample points. We defined the selected subset of the input objects as \textit{landmarks}. Landmarks, i.e., anchors, have been concurrently used with out-of-sample extensions to scale MDS and other embedding techniques~\cite{Silva_2002, Virtual_L}.

It remains to research what landmark selection method one should choose. Landmark selection may depend on the algorithm, nature of the input data and application area. We will be using the following methods in our experiments. 

\begin{itemize}

    \item Random selection :
    Randomly selects landmark points from the input dataset. This works well in practice~\cite{Silva_2004}, especially with large-scale data, since the computations are quick and cheap.    

    \item Farthest point sampling (FPS) is a standard method of selecting sources starting with an initial random point. Then it iteratively picks the farthest point from the already selected sources~\cite{KAMOUSI20161}. However, FPS accesses the entire distance matrix between all the available data points rather than a selected set of rows. Therefore, it is more computationally expensive than random selection.  
    
\end{itemize}

We recommend using random selection for quick approximations, which are cheap and effective for large-scale data. However, FPS has the advantage of being controllable when reproducible results are desired. We utilise the ideas of landmarks to propose an OSE technique that scales LSMDS for large-scale data.

\subsection{Optimisation Method}

We proposed an optimisation approach similar to the stress minimisation of the LSMDS algorithm to solve the OSE problem for LSMDS. However, we accomplish the OSE method with a different objective function because the original LSMDS objective involves all $N^2$ distances in the original set. In contrast, the OSE involves only the distances between a new object and the landmarks.

The proposed method is implemented as follows. For a given dataset of $N$ objects in a metric or non-metric space with dissimilarities ($\Delta_N=[\delta_{ij}]$), a configuration of {$\mathbf{X}=[x_1,x_2,..,x_N]$} points in a $K$-dimensional Euclidean space, where $x_i \in \mathbb{R}^K$ is constructed by applying the LSMDS algorithm. 

Then we choose a set of landmarks $L_s={l_1,...l_L}$, from the $N$ objects and the corresponding $\hat{l_1},...\hat{l_L}$ points from the $N$ pre-mapped points in the configuration space.

Finally, the mapping of the new point $y$ to $\mathbb{R}^K$ is calculated based on the pre-mapped positions of $L_s$ and the corresponding proximity information $\delta_{l_{i}y}$. Finding a new mapping position is considered a minimisation problem of stress similar to a standard LSMDS problem with $m$ points, where $m = L + 1$. However, only one point $y$ is movable among $m$ points. Hence the OSE of a new object $y$ is obtained by minimising the following objective function, 

\begin{equation} \label{eq:2}
    \hat{\sigma}(y) = {\sum_{i=1}^L}\big( \left\| \hat{l_i}-\hat{y}\right\|_2-\delta_{l_{i}y}\big)^2.
\end{equation}

The $\delta_{l_{i}y}$ represent the original dissimilarity value between $i^{th}$ landmark point and the new embedding point $y$ in the original space. The Euclidean distance between $i^{th}$ landmark and point $y$ in the existing configuration space in $K$ dimensions is given by~$\left\| \hat{l_i}-\hat{y}\right\|_2$. Thus $\hat{\sigma}(y)$ is used to match the original dissimilarities of the object $y$ to the landmarks into the Euclidean distances in the configuration space.

We seek to find a position of $\hat{y}$ that minimises $\sigma(y)$.  The proposed OSE algorithm simplifies the complicated non-linear optimisation problem to a small non-linear optimisation problem, such that $N$ points to $L + 1$ points non-linear optimisation problem, where $L \ll N$. Note, however, that it still requires the solution of a quadratic optimisation problem.

\subsection{Neural Network Method}

An ANN is a machine learning model that creates predictions based on existing data. An ANN consists of many simple processing elements, called neurons, each possibly having a small amount of local memory. These are organised into a sequence of layers linked by weights. The basic architecture of an ANN consists of:

\begin{itemize}
    \item Input layer: loads the input data from external sources to feed them into the model.
    \item Hidden layers: intermediate layers that do all the computations and feature extraction from the input data.
    \item Output layer: takes input from preceding hidden layers and comes to a final prediction based on the model’s learning.
\end{itemize}

ANNs can consist of several layers or many layers (deep neural networks). ANNs solve different types of problem areas, including classification, regression, multi-output or multi-class problems.

The networks are learned by processing the inputs and comparing the resulting outputs with the desired outcomes. The difference is then propagated to the network by adjusting its weights to minimise the error. It is an iterative process, and the weights are continually adjusted until the system converges to an optimum solution.

At an abstract level, an ANN is a function. Hence we can define the function $f_{\theta}:\Delta_y \rightarrow \hat{y}$ parameterised by $\theta$ which takes the input $\delta \in \mathbb{R}^L $ and produce the output $\hat{y} \in \mathbb{R}^K $~\cite{Ketkar2017}.

We formulate the OSE problem as a regression problem. Regression refers to predictive modelling problems that involve predicting a numeric value for a given input. Our task requires predicting more than one numerical value, usually known as the multi-output regression. Among the family of ANN models, we choose the multilayer perceptron (MLP), i.e., a multilayer, fully connected feed-forward network, to solve our out-of-sample regression problem.

As before, a configuration is obtained for $N$ points in $K$ dimensional space by applying the standard LSMDS algorithm. Then we choose a set of landmarks $L_s=l_1,...l_L$ from the $N$ pre-mapped points. 

Then the input data to the NN model is as follows. 

\begin{itemize}

    \item Input training data: The pairwise distances between the landmarks $L_s$ and the $N$ points measured in the original space. Hence the input data is a matrix $\Delta_{LN} \in \mathbb{R}^{L \times N}$. The  column vector of the $i^{th}$ sample in $\Delta_{LN}$ has the form $\delta^{i}=[\delta_{il_{1}},\delta_{il_{2}},\delta_{il_{3}},..,\delta_{il_{L}}] \in \mathbb{R}^{L}$, where $\delta_{il_{j}}$ is the distance between a landmarks and the $i^{th}$ point given $i \in {1,2,...,N}$ and $j \in {1,2,...,L}$. 

    \item Output training labels: The corresponding coordinate representation of $N$ points in the $K$-dimensional Euclidean space. The output labels of the $i^{th}$ sample is  $x^{i} \in \mathbb{R}^{K}$.

    \item Input testing data: The pairwise distances between the landmarks and a new point $y$ measured in the original space. Similar to the training data, the input testing data takes the form $\delta^y = [\delta_{yl1}, \delta_{yl_{2}}, \delta_{yl_{3}},...,\delta_{yl_{L}}] \in \mathbb{R}^{L}$. Testing data would usually include labels; however, we do not have labels for the out-of-sample points. 

\end{itemize}

Our MLP model is a by-now standard model with three hidden layers that use the ReLU activation function~\cite{Bo,Ketkar2017}. The size of the input layer is equal to the number of landmarks $L$, whereas the size of the output layer is $K$, representing the dimension of the output data. The sizes of the hidden layers are estimates of the intrinsic dimension of the previous layers. 

In each learning step, the ANN is presented with an input column vector $\delta^{i}$ and the label $x^{i}$. We will fit the model using the mean absolute error (MAE) loss function and the Adam version of stochastic gradient descent~\cite{Ketkar2017}. 

The MAE measures the average magnitude of absolute differences between $N$ label vectors $X={x_1,x_2,..,x_N}$ and predicted vectors $\hat{X}={\hat{x_1},\hat{x_2},...,\hat{x_N}}$, with the corresponding loss function; written as,

\begin{equation}\label{eq:3}
    Loss_{MAE}( X, \hat{X}) = \frac{1}{N} \sum_{i=1}^{N} || x_i - \hat{x_i}||,
\end{equation}

\noindent where, $|| x_i - \hat{x_i}||$ is the Euclidean distance between the label vector and the predicted vector of the $i^{th}$ sample of the input data.

Starting with initial random weights, MLP train the model to minimise the loss function by repeatedly updating these weights. The backpropagation algorithm is used in NN models to feed the loss backwards from the output layer to the previous layers, updating weights to decrease the loss. 

A comprehensive introduction to these topics is outside the scope of this thesis. See instead ~\cite{Ketkar2017, Goodfellow} for a general introduction to deep learning, and ~\cite{nature}, for backpropagation.

\noindent The out-of-sample LSMDS steps for both methods described above are as follows; (1) apply LSMDS on the training data, (2) choose landmarks from the training data, and  (3) apply any of the OSE techniques for the remaining data points based on the embedding results of the training data. We will discuss the results in Section~4.

\section{Experiment analysis}

We evaluated the performance of the two methods under various settings. Two main questions studied in these experiments are: (1) Are these proposed methods robust over multiple settings? And (2) Do any of the methods outperform the others in terms of scalability and accuracy? 

All algorithms are implemented in R and executed on a desktop with an Intel Core 5 Quad 2.3GHz, 16GB RAM, and MacOS Big Sur. We use Keras API and Tensorflow as the backend for developing our neural network model.

\subsection{Data Sets}

We examined the performance of our methods over synthetic datasets containing a set of strings, specifically a dataset of entity name strings. While entities can be any real-world object, they are mainly people in this context. These data are generated using the Geco tool in FEBRL~\cite{Geco}, where we can manipulate the data generation to significantly vary the size and characteristics (e.g., error rates) of the data. Each entity has a given name and a surname. The Geco tool allows us to create unique entity names as well as duplicate entries of entity names. We will be mainly using unique entity names in this work.

\subsection{Performance Evaluation}

We evaluate the results of the two OSE techniques based on an error metric $Err(m)$, similar to the stress function defined in~\autoref{eq:1}. It measures the distortion of the distances between new and existing points before and after the OSE transformation. The point error $PErr(y)$ measures the distances between a new point and the existing points before and after the OSE transformation. We defined them as follows;

The point error $PErr(y)$ of embedding a new point is given by, 

\begin{equation}\label{eq:4}
        PErr(y)=   \sum_{i=1}^{N}  (\delta_{iy} - ||x_i - \hat{y}||)^2 ,
\end{equation}

\noindent where the distance between the $i^{th}$ pre-mapped point and a new point $y$ in the original high dimensional space is given by $\delta_{iy}$, and the Euclidean distance between the same points in $K$ dimensions is given by $||x_i -\hat{y}||$.

The error $Err(m)$ of embedding $m$ new data points in a $K$ dimensional Euclidean space with a pre-mapped $N$ data points is given by,

\begin{equation}\label{eq:5}
        Err(m) =  \sum_{i,j=1}^{N,m} \frac{ (\delta_{iy_{j}} - ||x_i - \hat{y_j}||)^2} {\delta_{iy_{j}}},
\end{equation}

\noindent where $\delta_{ij}$ is the distance between the $i^{th}$ pre-mapped point and $y_{j}$ new point in the original high dimensional space, and $i \in 1,2,...,N$ and $y_{j} \in 1,2,...,m$. The Euclidean distance between the same points in $K$ dimensions is given by $||x_i-\hat{y_j}||$.

The error $Err(m)$ measures the embedding error of new points $m$ with respect to the $N$ points in the existing configuration. We calculated $Err(m)$ to compare the accuracy of the two OSE methods. Additionally, we used scatter plots to compare the point error values calculated for embedding each new point in $m$ using either OSE method. 

We also compared the efficiency of the methods using the computational time each method takes when processing out-of-sample data. All the CPU running times are measured in \textit{seconds} and denoted by RT.

\subsection{Choice of Parameters}

Several factors impact the performance of the proposed OSE methods; some of them (e.g., dimension($K$) and landmarks ($L$) ) are control parameters that we rely on to fine-tune the performance. The datasets determine the other important factors, such as dataset size.

In this section, We discuss the choices of $K$, $L$ parameters in our OSE implementations (and their rationale). In detail, we applied LSMDS and the proposed OSE techniques to a sample of data generated using Geco~\cite{Geco}. We chose 5000 name strings as the reference data points to create the initial configuration and separate 500 names as out-of-sample points. The reference data points and the out-of-sample points represent unique entities.

\textbf{K}: The dimension of the Euclidean space ($K$) plays an essential role in the performance of the LSMDS algorithm. A good value of $K$ should differentiate similar objects from dissimilar ones by approximating the actual distances between them. Embedding strings using LSMDS has been studied before in the context of entity resolution. We set the $K=7$, as recommended in~\cite{herath2020simulating}, where the authors have investigated the trade-off between the \textit{stress vs dimension}.

\textbf{L}: The number of landmarks is a dominating factor that directly affects the effectiveness and efficiency of the proposed OSE techniques. Landmarks reduce the computational overhead of distance calculations. Since only a fraction of the initial data is used to map a new point to the configuration space, it also reduces the total running time of the OSE process.


We applied two landmarks selection methods: FPS and random selection. The following results are based on the FPS. The number of landmarks to be used remain to investigate in what follows. 

The use of a few landmarks only requires a few distance calculations. Hence, the method becomes efficient for large-scale data processing. In contrast, fewer landmarks might produce a less accurate approximation of the mapping positions for new data. We have to consider this trade-off when choosing an adequate number of landmarks appropriate for our data. The following experiments measure the effect of landmarks on the performance of the proposed OSE techniques. The performance is measured based on the accuracy of the distance approximation and the average RT of mapping a single point using either of the methods.


\subsubsection{Comparing the Total Error}

First, we applied LSMDS to the selected 5000 strings of reference points. Then the remaining 500 strings (out-of-sample points) are mapped using the proposed OSE methods, only measuring the distances to the landmarks. We measured the total error ($Err(m)$) generated by mapping the out-of-sample data in the configuration space for different instances varying $L$. We obtained two sets of results (in terms of total error) for both OSE techniques by keeping the $K=7$ fixed and changing $L$.

\begin{figure}[ht!] 
\centering
\includegraphics [height=1.8in, width=3.5in]{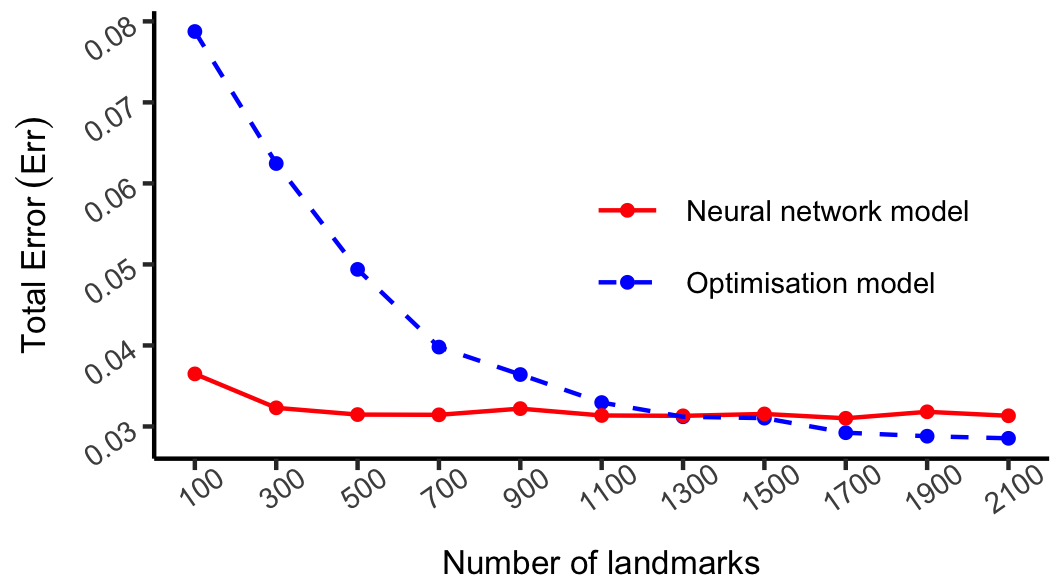}
\caption{A comparison between the proposed out-of-sample techniques: neural network model and the optimisation method. The total error ($Err(m)$ in \autoref{eq:5}) represents the distance distortion created through the mapping process when mapping out-of-sample data from the high dimensional space to the Euclidean space. The $Err_{o}(m)$ decreases rapidly for the optimisation method. In contrast, the total error fluctuates in small amounts for the neural network when the number of landmarks increases in each instance.}
\label{fig:1}
\end{figure}

\autoref{fig:1} compares the two OSE techniques; the NN model and the optimisation method. We utilised the same set of landmarks when applying both methods in each instance. Let $Err_{o}(m)$ and $Err_{nn}(m)$ be the total errors produced by the optimisation method and the NN model, respectively. 

The optimisation method reduces the $Err_{o}(m)$ drastically until the number of landmarks is 1000. Then $Err_{o}(m)$ reduction gradually slows down. The $Err_{o}(m)$ approaches a non zero asymptotic, and it depicts that the increasing number of landmarks increases the accuracy of the OSE. 

Small $L$ tends to produce higher $Err_{o}(m)$ values in the optimisation method since only a few distance calculations are used in the approximations. If $L$ is set very high ($L=2000$), the $Err_{o}(m)$ decreases as we have more distances to approximate the mapping of the new data points. However, having a large number of landmarks can increase the processing time of the OSE. Thus $L$ should be set neither too high nor too low, suggesting landmarks between 1000 t0 1500 gives a reasonable $Err_{o}(m)$ for the overall approximation of the given data set when applied the optimisation approach. 

In contrast, the NN model only shows a significant improvement when landmarks increase from 100 to 300. After that, $Err_{nn}(m)$ does not significantly reduce with the increasing number of landmarks used in the model training. The results suggested that the neural network OSE method performs well with fewer $L$ and sufficient training data. In contrast, the optimisation method requires more landmarks to achieve the same level of accuracy as the NN model. While both methods perform similarly around 1100-1500 landmarks, the NN model works very efficiently around 100-300 landmarks with a slightly reduced accuracy.

\subsubsection{Comparing the Point Errors}

\begin{figure}[ht!] 
  \begin{subfigure}{6.9cm}
    \centering\includegraphics[width=6.9cm, height=2.4in]{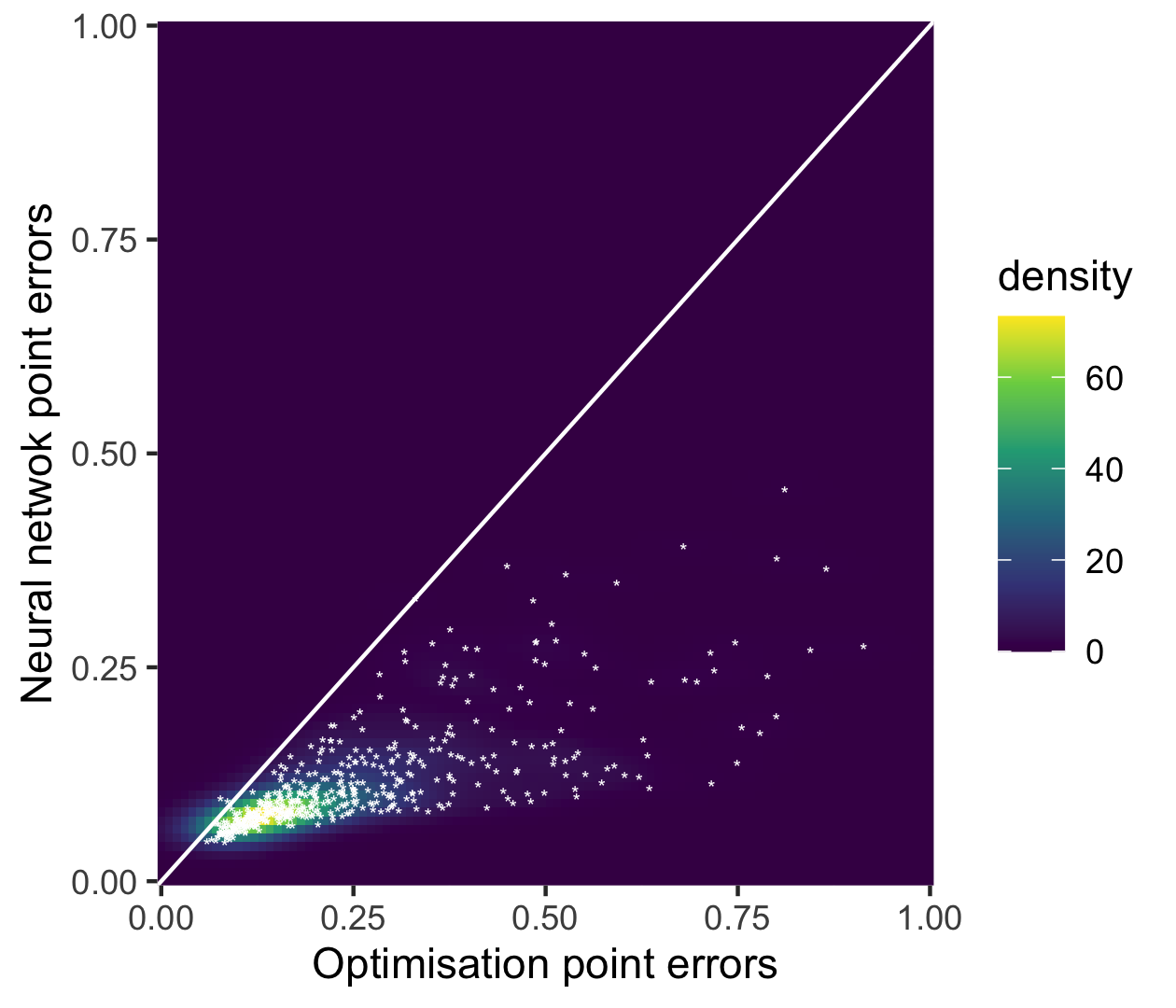}
    \caption{L=100}
  \end{subfigure}
  \begin{subfigure}{6.9cm}
    \centering\includegraphics[width=6.9cm,height=2.4in]{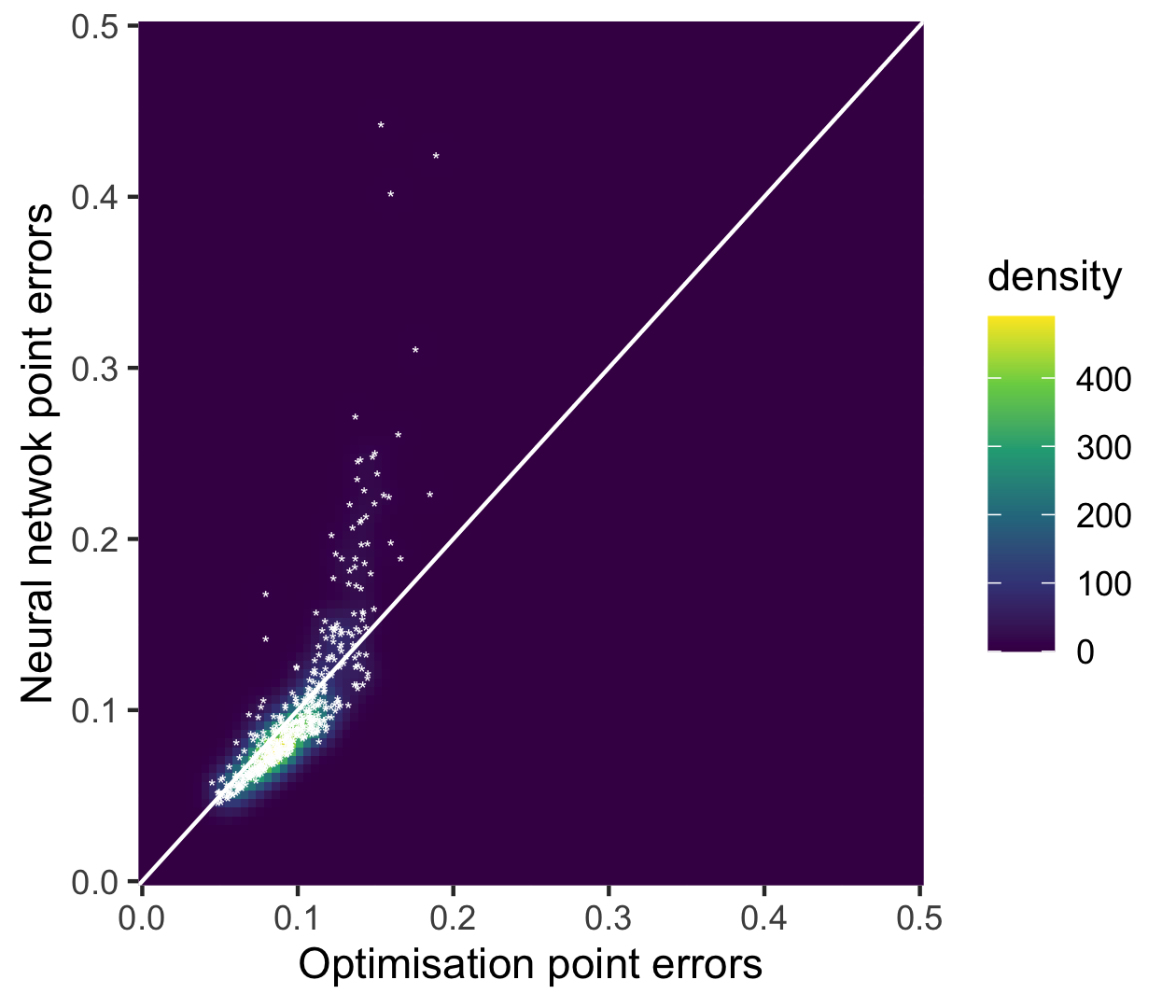}
    \caption{L=1500}
  \end{subfigure}
  \caption{The comparisons between the point errors of embedding out-of-sample data using the neural network and the optimisation methods. The NN model is better for the points below the $y=x$ line, and optimisation is better for the points above the line. a) The point errors produced by both OSE approaches with 100 landmarks. The NN model produces small errors compared to the optimisation b) The point errors created by both OSE approaches with 1500 landmarks. Both methods make similar errors with many landmarks, and the magnitude of the point errors are much smaller.}
  \label{fig:2}
\end{figure}

Next, we looked at the error of distortion in distance approximation for each out-of-sample point separately. The experiment settings are the same as before, and instead of the total error, we calculated the distance approximation errors separately for each out-of-sample mapping. We applied the OSE methods to map each new string to the existing configuration space and calculated $PErr(y)$ (\autoref{eq:4}) for every single mapping. We obtained two sets of results produced by the two methods. \autoref{fig:2} compares those point error values obtained for the underlying out-of-sample data points. Here we have normalised the $PErr(y)$ by $\delta_{iy_{j}}$ which is the sum of the dissimilarities between all the new points and the existing points in the original space.

Ideally, we want point errors to be zero for each OSE point. However, they cannot be zeros since our methods only approximate the distances between new points and landmarks, excluding the distances between the remaining points and the new points in the embedding calculations. On the other hand, no exact solution exists for our data since the embeddings are invariant to position, and the methods only approximate the relative distances. The point errors are calculated considering the distances from a new point to all the existing points to measure the error of distance approximation produced by the proposed OSE techniques. Hence, our methods tolerate a small amount of error in mapping a new data point to provide a more efficient solution at the expense of slight accuracy loss.

\autoref{fig:2} describes how the point errors change with the number of landmarks when the proposed OSE techniques are applied for new data points. We have measured the point errors of embedding 500 out-of-sample entity name strings in a $7$-dimensional Euclidean space. The plots are drawn only for $L= 100, 1500$, showing significant improvements with more landmarks.

\autoref{fig:2}-(a) illustrates $PErr(y)$ comparisons of the mapping based on $L=100$ landmarks. All the point errors produced by the NN model are less than 0.5, whereas the point errors produced by the optimisation method are less than 0.75. It depicts that the NN model can approximate the mapping of a new point with few landmarks producing a small error. In contrast, the optimisation method creates comparatively large point errors in the mapping with few landmarks.

\autoref{fig:2}-(b) illustrates $PErr(y)$ comparisons of the mapping based on $L=1500$ landmarks. Increasing the number of $L$ has improved the accuracy of both methods. All the point error values produced by the optimisation method are below 0.25. Similarly, 95\% of the point error values produced by the NN model lies below 0.25. Hence, with a sufficient amount of landmarks, we can improve the accuracy of both methods where the optimisation method remains as good as the NN model.

\begin{figure}[ht!] 
  \begin{subfigure}{6.9cm}
    \centering\includegraphics[width=6.9cm, height=2.4in]{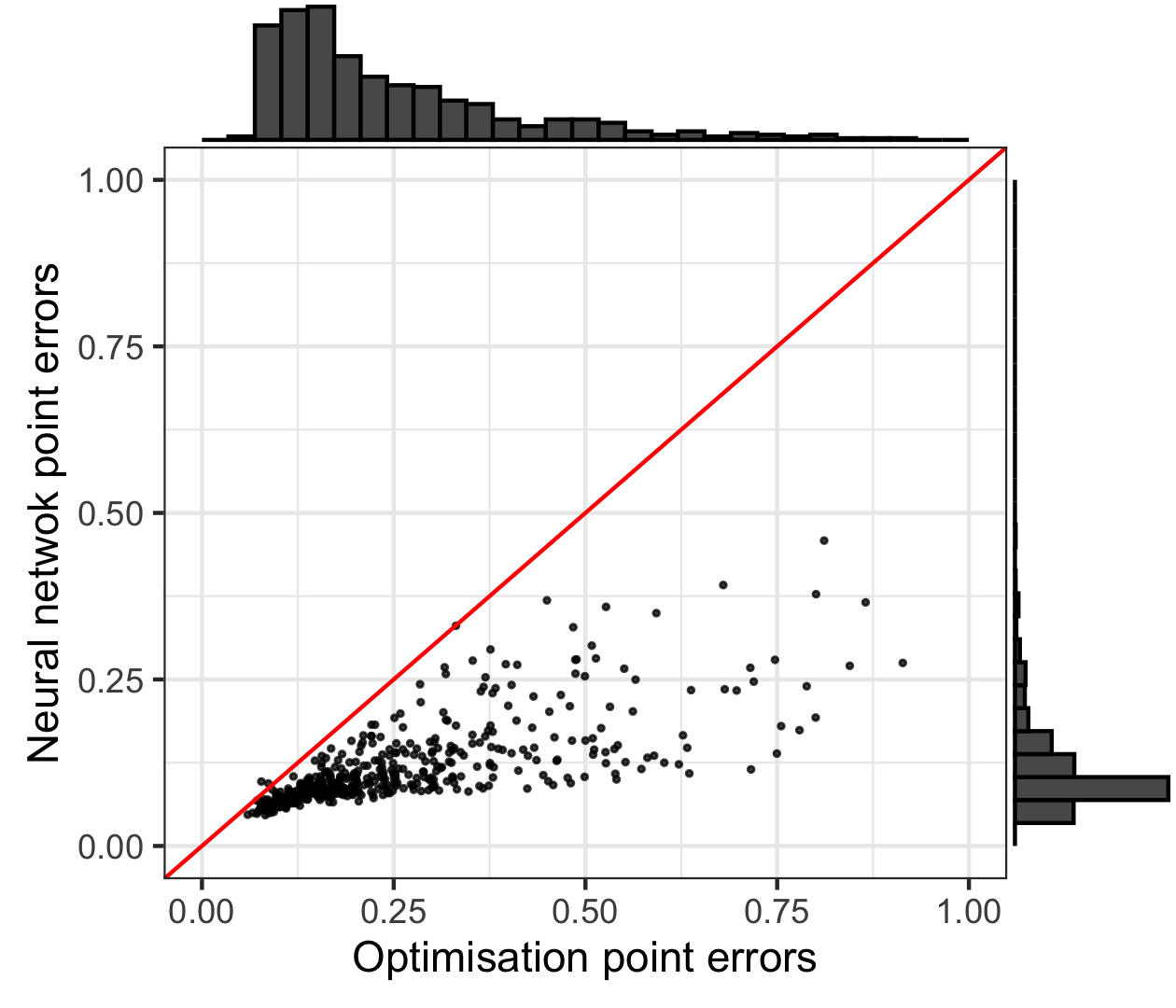}
    \caption{L=100}
  \end{subfigure}
  \begin{subfigure}{6.9cm}
    \centering\includegraphics[width=6.9cm,height=2.4in]{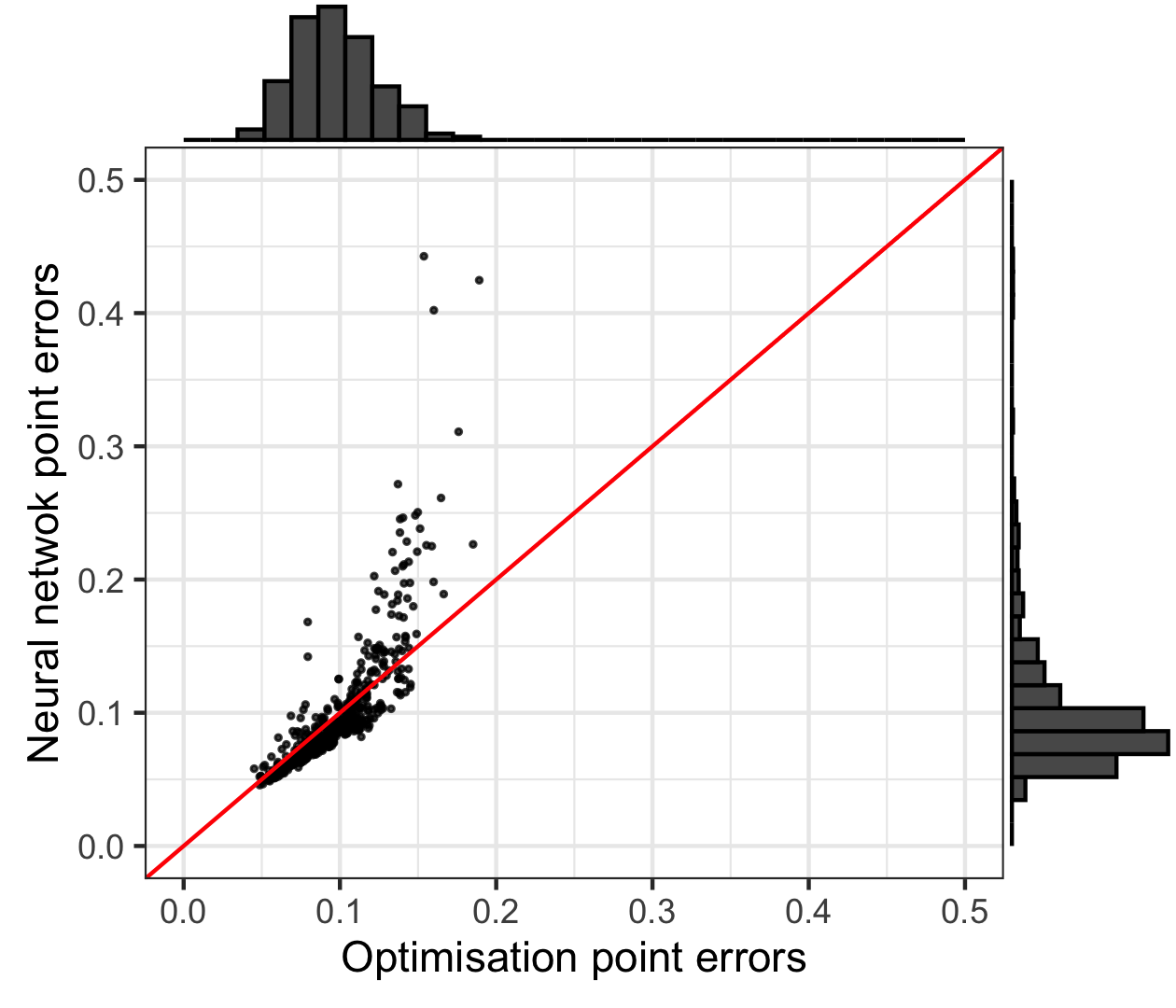}
    \caption{L=1500}
  \end{subfigure}
  \caption{The comparisons between the point errors of mapping out-of-sample points and their distributions. The neural network model and the optimisation method are applied to map out-of-sample points into a $7$-dimensional Euclidean space.
  a) The point errors and their distributions created by both out-of-sample approaches with 100 landmarks. The point error distribution of the neural network results shows a small spread indicating small variance, whereas the optimisation results show a comparatively wider spread indicating high variance. 
  b) The point errors and their distributions created by both out-of-sample approaches with 1500 landmarks. Both distributions have similar shapes and a smaller spread indicating small variances when the number of landmarks increases to 1500.}
  \label{fig:3}
\end{figure}

Next, we looked at the $PErr(y)$ distribution produced by both approaches. \autoref{fig:3} compares the point errors and their distributions of the two OSE methods. These figures extend the $PErr(y)$ values shown in \autoref{fig:2} to their distributions. When $L=100$, the error distribution of the neural network results shows a small spread indicating small variance, whereas the optimisation results show a comparatively wider spread indicating slightly higher variance. In contrast, both distributions have similar shapes and a smaller spread indicating small variances when $L$ increases to $1500$. Nevertheless, using many landmarks in the embedding can be inefficient for large-scale data. In the following experiment, we explain the running time complexities of the proposed methods.

\subsubsection{Comparing the Average Running Time}

We evaluated the efficiency of the methods measuring the RT of mapping a new point in the existing configuration space. The number of landmarks $L$ directly impacts the running time (RT) of the proposed OSE methods since each new point is mapped to the existing configuration space using landmarks. Both methods map a single out-of-sample point at a time.

\begin{figure}[ht!] 
\centering
\includegraphics [height=1.8in, width=3.4in]{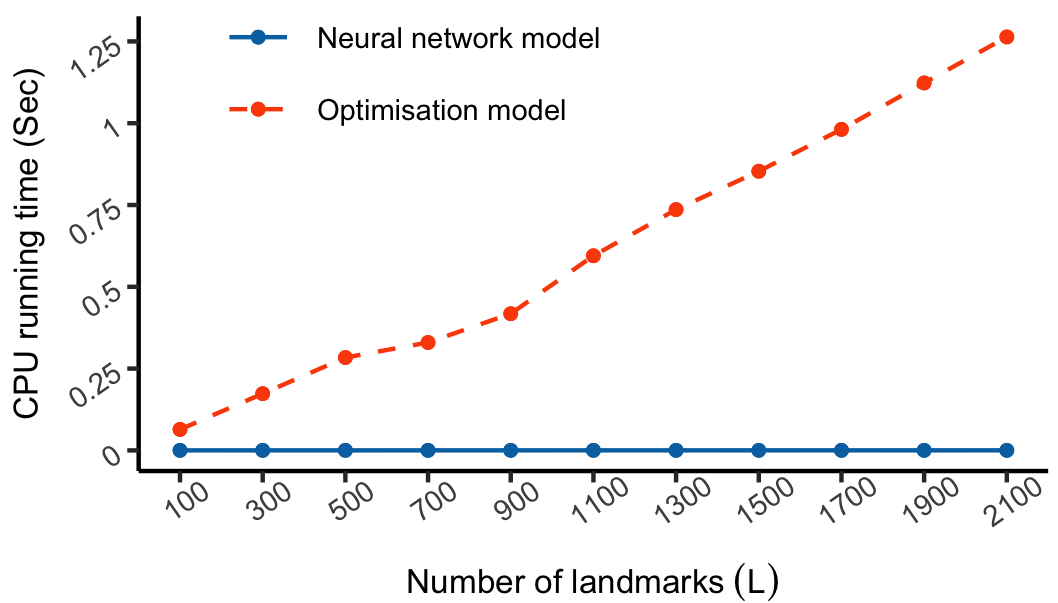}
\caption{A comparison of the average running time of mapping a single query in an existing configuration space using the proposed out of sample methods: neural network and optimisation. Each instance represents a different number of landmarks used in the mapping. The running time increases linearly with the number of landmarks for both methods, although only the optimisation method shows a significant increase in the RT rate per landmark.
}
\label{fig:4}
\end{figure}

\autoref{fig:4} compares the average running time (RT) of mapping an out-of-sample point in the configuration space for both methods, varying the numbers of landmarks. Increasing $L$ increases the average mapping time of a single point linearly in the optimisation method. The average mapping time increases linearly with $L$ for the neural network model, which is not visible due to the scaling. When $L=100$, the average RT of mapping an out-of-sample point only takes $0.013 \times 10^{-2}$, and with $L=2100$ the average RT is $0.2 \times 10^{-2}$. The total rates of change in RT with respect to landmarks are $10\%$ and $7.5 \times 10^{-4}\%$ for the optimisation and NN models, respectively. 

The results show that the NN model is faster than the optimisation method regardless of the number of landmarks used in the mapping. 
Both methods spend less RT around $100$ landmarks and increase linearly with $L$. The RT grows gradually (small gradient) for the NN model and drastically (large gradient) for the optimisation model.  However, a small number of landmarks can reduce the accuracy of both methods. This was a significant observation in our previous experiments, as illustrated in \autoref{fig:1}.

These results also support our earlier remarks on the neural network that emphasise that the model can be very fast with a small cost of accuracy. We also discussed that both methods have similar accuracy levels between 1000 to 1500 landmarks. However, the NN model is on average $3.8 \times 10^{3}$ faster than the optimisation method around $L={1000,1500}$. Hence, we can achieve high accuracy and efficiency using the NN model with fewer resources than the optimisation method. Finally, we conclude that the NN model is a highly accurate and efficient method and an excellent substitution for the optimisation OSE method. With less than $1000$ landmarks, the NN model can map an out-of-sample point within $1.7 \times 10^{-4}$ seconds for our data. Hence, the average running time of mapping an out-of-sample point into an existing configuration space takes less than a millisecond. 


\section{Discussion}

The NN model maps a new point into an existing configuration space within less than a millisecond, achieving a good approximation for the actual distances. 

In contrast, the optimisation method requires many landmarks to achieve the same level of accuracy. Many landmarks are computationally expensive since they increase the embedding RT. Hence, there is a trade-off between the accuracy and efficiency in the optimisation approach, which depends on $L$. However, this method directly predicts the mapping coordinates of a new object in an existing configuration space using distances to the landmarks. In contrast, the NN model requires training to build the model, which takes time. However, once the model has been trained, the OSE of new data is immediate. Since our NN model is a simple MLP with three hidden layers, it only takes approximately 1.2 seconds to train the model for our data.

The optimisation method requires an initial guess (or a ``starting point") for the predicting value. The choice of a starting point determines how quickly the algorithm converges to a solution. We choose a vector of all $0$s as the initial guess of mapping a new object in the configuration space in our experiments. Improving the initial guess is non-trivial since the initial configuration of the configuration space provided by the LSMDS is not invariant to orientation or position. Hence, one of the reasons for the optimisation method to provide less accuracy can be the sensitivity of the algorithm to the initial guess.

\section{Conclusion}

In this work, we have proposed and tested two out-of-sample methods that extend the LSMDS algorithm to large-scale datasets. The proposed techniques are based on an optimisation approach and an artificial neural network. Both methods proved fast running time and scalability along with the data size. However, the NN model is faster than the optimisation method. We combined the ideas of landmarks and OSE to address the scalable issues in LSMDS. The basic idea is first to embed a subsample of the input data by applying LSMDS and then map the remaining data as out-of-sample points. Hence, we can apply the NN model to embed large-scale data sets into configuration space. Our method can also map previously unseen data into an existing configuration space. 

Many directions are ahead for future work. First, we plan to extend the out-of-sample method to be parallel. Future research also includes applying these ideas to different areas, including entity resolution, DNA visualisation.

\bibliographystyle{ieeetr}
\bibliography{ref}

\end{document}